\documentclass[lettersize,journal]{IEEEtran}
\usepackage{amsmath,amsfonts}
\usepackage{algorithmic}
\usepackage{algorithm}
\usepackage{array}
\usepackage[caption=false,font=normalsize,labelfont=sf,textfont=sf]{subfig}
\usepackage{textcomp}
\usepackage{stfloats}
\usepackage{url}
\usepackage{verbatim}
\usepackage{graphicx}
\usepackage{cite}
\def \cls{[\textsc{Cls}]}
\usepackage{multirow}
\usepackage[table]{xcolor}
\usepackage{colortbl}
\usepackage{makecell}
\begin{document}

\title{Rethinking the Global Knowledge of CLIP in \\Training-Free Open-Vocabulary Semantic Segmentation}

\author{Jingyun Wang, Cilin Yan, Guoliang Kang
\thanks{Jingyun Wang, Cilin Yan, and Guoliang Kang are with Beihang University, Beijing, China. E-mail: \{wangjingyun0730, clyanhh, kgl.prml\}@gmail.com.}

\thanks{(Corresponding author: Guoliang Kang)}%
\thanks{This paper has supplementary downloadable material available at http://ieeexplore.ieee.org., provided by the author. The material includes more ablations and visualizations. This material is 4 pages in size.}
\thanks{Manuscript received April 19, 2021; revised August 16, 2021.}
}

\markboth{Journal of \LaTeX\ Class Files,~Vol.~14, No.~8, August~2021}%
{Shell \MakeLowercase{\textit{et al.}}: A Sample Article Using IEEEtran.cls for IEEE Journals}


\maketitle

\begin{abstract}
Recent works modify CLIP to perform open-vocabulary semantic segmentation in a training-free manner (TF-OVSS).
In vanilla CLIP, patch-wise image representations mainly encode homogeneous image-level properties, 
which hinders the application of CLIP to the dense prediction task.
Previous TF-OVSS works sacrifice globality 
to enhance the locality of CLIP features, 
by making each patch mainly attend to itself or its neighboring patches within a narrow local window.
With their modifications, the ability of CLIP to aggregate global context information is largely weakened.
Differently, in this paper, we rethink the global knowledge encoded by CLIP and propose GCLIP to answer how to extract and utilize beneficial global knowledge of CLIP for TF-OVSS.
As the representation of each patch is finally determined by the attention weights and the Value embeddings, we propose to reshape the last-block attention and Value embeddings to aggregate useful global context into final features. 
Firstly, we aim to equip the last-block attention with image-level properties while not introducing homogeneous attention patterns across patches. 
To realize the goal, we fuse the attention from the global-token emerging blocks with the Query-Query attention.
Secondly, we aim to make Value embeddings of the last-block attention module more semantically correlated. To realize this, we design a novel channel suppression strategy.
Extensive experiments on five standard benchmarks demonstrate that our method consistently outperforms previous state-of-the-arts.
\end{abstract}


\begin{figure*}
  \centering
  \includegraphics[width=1\linewidth]{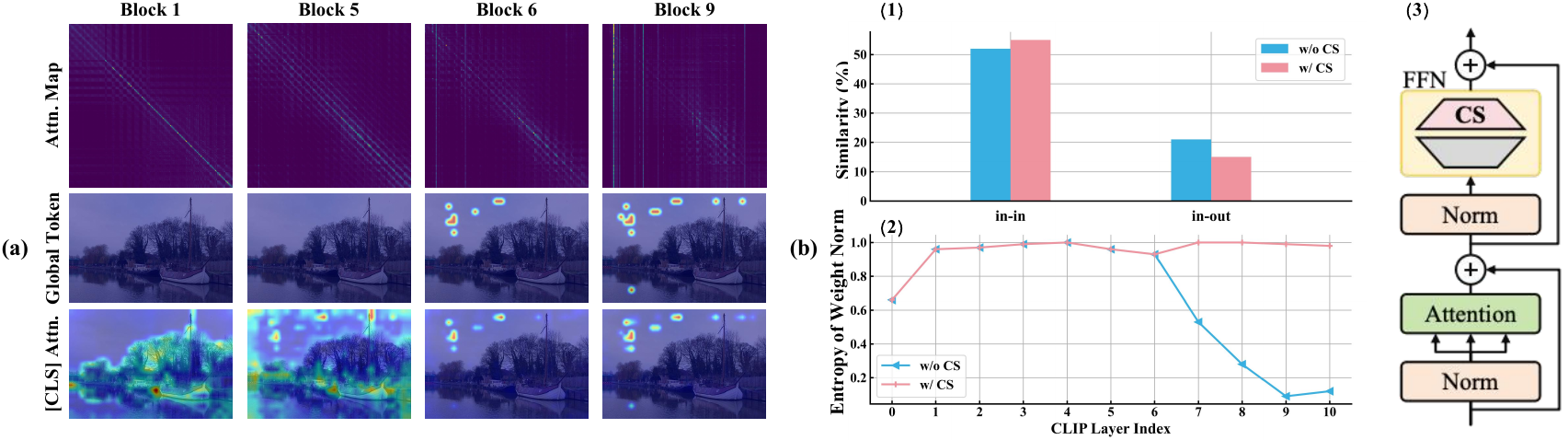}
  \caption{Experiments with CLIP ViT-B/16.
  (a) \textbf{Emergence of global tokens (best viewed in color).} Global tokens (highlight stripes in Line 1) start to emerge from the attention map of block 6. Comparing the attention maps after block 6, we observe the attention pattern of global tokens aligns well with that of the \cls{} token (Line 2\&3). (b) \textbf{Channel Suppression (CS).} We observe the entropy of weight norms decreases abnormally from block 7 in (2). With CS on the abnormal weight norm of the second fully-connected layer of FFN in a Transformer block (See (3)), we enhance the semantic correlation by making value embeddings of patches within the same semantic mask become more similar (``in-in'') but those from different masks become more dissimilar (``in-out'').}

\label{introduction_fig}
\end{figure*}

\section{Introduction}
\label{sec:intro}
\IEEEPARstart{S}{emantic} segmentation aims to assign a semantic label to each pixel within an image. 
With the rise of deep learning, the performance of semantic segmentation~\cite{FCN, Deeplabv1, Deeplabv2, PSPNet, Mask2Former, MaskFormer, TMM1,TMM2,TMM3} has been dramatically improved, but still relies on closed-set training covering a limited number of categories.
In the real world, there are a large number of open-vocabulary classes that are not seen during training, and the closed-set semantic segmentation methods may not be able to make predictions for them.
To deal with open-vocabulary semantic segmentation (OVSS) problem, many methods~\cite{GroupViT, TCL, CoCu, ODISE, OVSeg, OpenSeg, SimSeg, LSeg, CATSeg} have been developed and exhibit superior generalization ability to unseen categories. 
However, most of OVSS methods still heavily rely on time-consuming training with large-scale image-caption pairs or class-agnostic masks, which hinders the application of OVSS methods in practice.

Recent works modify large-scale vision-language pre-trained model  CLIP~\cite{CLIP} to perform OVSS in a \textit{training-free} manner. 
Though CLIP demonstrates superior zero-shot performance for image classification task, it cannot be directly applied to OVSS, as the patch-wise representation of CLIP tends to encode homogeneous image-level properties, 
hindering pixel-level prediction. 
Previous methods for TF-OVSS~\cite{MaskCLIP, ClearCLIP, CLIPSurgery, CLIPtrase, SCLIP} view global knowledge of CLIP as harmful for segmentation.
They modify the attention mechanism in the final block of CLIP, which encourages each patch to primarily focus on itself or the neighboring patches within a narrow local window.
Though image features are more distinct across patches, the CLIP's ability to aggregate global context information is significantly weakened.
As a result, the segmentation performance of these works is largely constrained.

In this paper, we rethink the global knowledge encoded by CLIP and propose GCLIP.
While the importance of global information is recognized in conventional semantic segmentation practice~\cite{PSPNet,Deeplabv1}, it is nontrivial to utilize the global feature of CLIP for segmentation and thus previous TF-OVSS works choose to sacrifice globality to enhance the locality of CLIP features.
We aim to bridge this gap by answering how to extract and utilize beneficial global knowledge of CLIP, under the specific TF-OVSS setting.
As the representation of each patch is finally determined by the attention weights and the Value embeddings, we make modifications to the last-block attention and Value embeddings respectively to aggregate useful global context information into final features. 
Firstly, we propose an Attention Map Fusion strategy (AMF) to emphasize global knowledge by reshaping last-block attention.
As shown in Figure~\ref{introduction_fig} (a), we observe that \textit{global tokens} exist in deeper blocks of CLIP. 
The term ``global token'' means a specific patch is important (\emph{i.e.,} corresponding Query-Key attention weight is high) for all the other patches, which is a consequence of CLIP's pre-training.
Interestingly, we find the attention pattern of global token aligns well with that of \cls{} token (see Figure~\ref{introduction_fig} (a)),
which indicates those global tokens may encode the image-level properties as [CLS] token. 
Based on such observations, we propose AMF to average the attention maps from global-token emerging blocks and the final-block Query-Query attention to form a new final-block attention.
Therefore, through AMF, we emphasize the global knowledge encoded by global tokens. 

Secondly, we propose a Channel Suppression (CS) strategy to make last-block Value embeddings more semantically correlated, which means the similarity between Value embeddings can reflect their semantic correlation.
In vanilla CLIP, we observe that the same channel in different Value embeddings has super-large activation, rendering Value embeddings across patches unexpectedly similar.
This is due to an abnormal phenomenon that exists in the weights of the second fully-connected layer of FFN in a Transformer block.
In detail, the weight norm corresponding to some specific output channels becomes unexpectedly larger than the weight norm of other channels, which can be reflected by the entropy of those weight norms (Figure~\ref{introduction_fig} (b)(2)).
Thus, we propose to suppress the abnormal weight norm of FFN (see Figure~\ref{introduction_fig} (b)(3)) so that the semantic correlation of Value embeddings can be enhanced. 
With CS, as shown in Figure~\ref{introduction_fig} (b)(1), we observe the Value embeddings of patches within the same semantic mask become more similar (see ``in-in'' comparison) while those from different masks become more dissimilar (see ``in-out'' comparison).
Since the representation of each patch is finally determined by the attention weights and the Value embeddings, we finally generate more semantically correlated patch-level image features while also absorbing global context.

We conduct extensive experiments on five standard semantic segmentation benchmarks, including PASCAL VOC~\cite{PASCAL_VOC}, PASCAL Context~\cite{PASCAL_Context}, ADE20K~\cite{ADE20K}, Cityscapes~\cite{cityscapes} and COCO Stuff~\cite{cocostuff}. 
Experiment results demonstrate that GCLIP consistently outperforms previous state-of-the-arts.
Notably, on Cityscapes, our method outperforms 
ClearCLIP~\cite{ClearCLIP} by 3.7\% mIoU. 
Extensive ablation studies further verify the effectiveness of each design.

In a nutshell, our contributions are summarized as 
\begin{itemize}
\item 
We propose an Attention Map Fusion strategy (AMF) to emphasize the global knowledge encoded by global tokens via reshaping the last-block attention.

\item 
We propose a Channel Suppression strategy (CS) to make last-block Value embeddings more semantically correlated.

\item 
We conduct extensive experiments on various segmentation benchmarks under the training-free open-vocabulary setting. Experiment results show that GCLIP outperforms previous state-of-the-arts. 
\end{itemize}

\section{Related Work}
\label{sec:related_work}

\noindent \textbf{Pre-trained vision-language models}
Pre-trained vision-language models (VLMs)~\cite{Uniter, Virtex, Unicoder-vl, AlignBeforeUse, Hero} have experienced rapid development, thanks to the abundant large-scale image-text pairs accessible on the Internet.
Recently, CLIP~\cite{CLIP}, ALIGN~\cite{ALIGN} and Slip~\cite{Slip} have made great progress on learning visual and textual representations jointly by using contrastive learning. 
Among these, CLIP trained on WIT-400M exhibits robust zero-shot capability for the image classification task, due to its image-level alignment with text.
However, directly applying CLIP to dense prediction tasks, such as object detection and semantic segmentation, results in suboptimal performance.
A series of methods~\cite{DetPro,UniDetector, MaskCLIP,GroupViT,TCL,ReCLIP} have successfully adapted CLIP for various downstream tasks, and this paper specifically addresses the adaptation of CLIP for the task of training-free open-vocabulary semantic segmentation.

\noindent \textbf{Open-vocabulary semantic segmentation (OVSS)}
OVSS refers to segmenting an image with arbitrary categories under the guidance of a textual description.
Among these, fully supervised OVSS~\cite{OpenSeg,LSeg,ODISE,OVSeg,CATSeg} methods still rely on high-quality pixel-level annotated masks.
Usually, they generate mask proposals by an extra mask generator, \emph{e.g.}, Mask2Former~\cite{Mask2Former}, and further align the visual embeddings with the textual features.
Most methods extract visual features by CLIP, while ODISE leverages the internal representations of pre-trained Diffusion models~\cite{Diffusion}.
Methods for fully supervised OVSS usually train on a large-scale dataset equipped with fully annotated masks, like COCO Stuff~\cite{cocostuff}, and directly perform zero-shot inference on other datasets that may contain unseen categories during the training process.
There also exists a set of OVSS methods~\cite{CoDe,GroupViT,viewco,CoCu},which mainly exploit large-scale image-caption pairs, such as CC12M~\cite{cc12m} and YFCC~\cite{yfcc}, for training.
For example, GroupViT~\cite{GroupViT} introduces grouping tokens into the vision transformer and conducts hierarchical clustering for segmentation.
It finally obtains an image-level feature, which is then aligned with textual features by contrastive learning loss.

\noindent \textbf{Training-free open-vocabulary semantic segmentation}
Methods for TF-OVSS~\cite{MaskCLIP,CLIPSurgery,SCLIP,CLIPtrase,ClearCLIP} perform OVSS without any training.
Existing CLIP-based TF-OVSS works explore to enhance the distinction across the patch-wise visual features from CLIP mainly by modifying the attention mechanism in its final block, which forces each patch to primarily focus on itself and its neighbors in a narrow local window.
For example, CLIPSurgery~\cite{CLIPSurgery} and GEM~\cite{GEM} replace the conventional Query-Key attention with Value-Value attention. During forward, 
they additionally align the new self-attention input with the vanilla input to avoid deviation accumulation.
However, with the proposed self-self attention, the ability of CLIP to aggregate global context information, which is known to be useful for distinguishing confusing categories, is weakened.
Our proposed GCLIP in this paper belongs to the category of TF-OVSS methods, and we mainly compare with the methods under the same setting for fairness.
We observe some contemporary TF-OVSS works~\cite{ResCLIP,Corrclip}.
Among these, ResCLIP~\cite{ResCLIP} shares a similar attention map fusion strategy with ours but differs in motivation and specific technical design (\emph{e.g.,} compared to ResCLIP, we design a more systematic way to determine which layers' attention maps should be fused, which is more practical and extendable).
Corrclip~\cite{Corrclip} adopts an additional large-scale pre-trained model, SAM~\cite{SAM}, as a mask generator to reconstruct patch correlations in CLIP and refine the segmentation mask.
\begin{figure*}
  \centering
\includegraphics[width=1\linewidth]{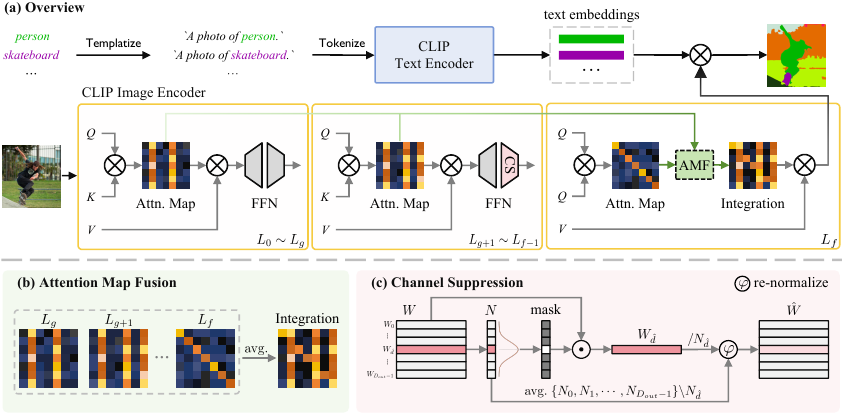}
  \caption{\textbf{Method Overview.} 
  (a) \textbf{Overview.} In this paper, we propose a new framework GCLIP, consisting of Attention Map Fusion (AMF) and Channel Suppression (CS), for Training-Free Open-Vocabulary Semantic Segmentation. 
  (b) \textbf{Attention Map Fusion.} We fuse the attentions of early global-token emerging blocks ($L_g$,$L_{g+1}$, $\cdots$) with the Query-Query attention of the last-block ($L_{f}$) to emphasize the effect of global knowledge.
  (c) \textbf{Channel Suppression.} We suppress the weight norm of the specific output channel $\hat{d}$ of FFN by a re-nomalizing operation $\varphi$ as depicted in Eq.~(\ref{formula:renormalize}) to enhance the semantic correlation of Value embeddings.}
\label{method_fig}
\end{figure*}

\section{Method}
\label{sec:method}

\noindent \textbf{Overview}
In this work, we propose GCLIP, a new framework for Training-Free Open-Vocabulary Semantic Segmentation (TF-OVSS).
The general framework of our method is illustrated in Figure~\ref{method_fig}.
The textual input is formed by filling the category name in the manually designed prompt, \emph{e.g.}, 
``a photo of a \#classname''.  
Passing the textual input into the text encoder of CLIP, we obtain the text embeddings $Z_{\text{text}}$.
Previous work, ClearCLIP~\cite{ClearCLIP} for TF-OVSS, enhances the locality across patches but harms the capability of CLIP to exploit global context (Sec.~\ref{3.1baseline}).
Based on ClearCLIP, we propose GCLIP with two simple yet effective modifications to the last-block attention and Value embeddings respectively to mine the beneficial global knowledge of CLIP for TF-OVSS.
Firstly, we propose an Attention Map Fusion strategy (AMF) to emphasize the global knowledge encoded by global tokens via reshaping the last-block attention (Sec.~\ref{3.2attnmap}).
Secondly, we propose a Channel Suppression strategy (CS) to make last-block Value embeddings more semantically correlated (Sec.~\ref{3.3wn}).
We forward the visual input $I\in\mathbb{R}^{3 \times H\times W}$ through the visual encoder of GCLIP.
Since the representation of each patch is finally determined by the attention weights and the Value embeddings, we can finally generate more semantically correlated patch-level image features $Z_{\text{GCLIP}}$ while also absorbing global context information.
By comparing the similarity between $Z_{\text{GCLIP}}$ and $Z_{\text{text}}$, we generate a logit map and further predict the segmentation mask by $\text{argmax}$ operation on the logit map.

\subsection{Baseline}
\label{3.1baseline}
In this paper, we adopt ClearCLIP~\cite{ClearCLIP} as our baseline model.
ClearCLIP modifies the final block $L_f$ of CLIP to enhance the distinctness of patch-wise representations for TF-OVSS.
In detail, ClearCLIP alters the last-block Query-Key attention to Query-Query attention, which enables each patch to mainly focus on itself.
Besides, ClearCLIP discards the residual outputs from other blocks, as they introduce global characteristics that are homogeneous across patches and harm the patch-wise distinction.
Additionally, since the removal of residual connection significantly changes the input to the last-block FFN, ClearCLIP further discards the last-block FFN to mitigate the negative effect. 
As a result, ClearCLIP simply adopts the output of the last-block Query-Query attention module for vision-language inference: 
\begin{equation}
Z_{\text{ClearCLIP}} = \text{Proj}({A}_{f}^{qq}\cdot v),
\end{equation}
where $\text{Proj}$ refers to the output projection in the multi-head self-attention module, 
${A}_{f}^{qq}$ and $v$ refer to the Query-Query attention map and Value embeddings from the final block $L_f$.

Although ClearCLIP enhances the distinction of the image features across the patches, it significantly weakens the capability of CLIP to aggregate global context information, which may provide a global view of the image and benefit in distinguishing confusing categories in the dense prediction task.
For example, in Figure~\ref{visualization_fig}, due to insufficient global context information, ClearCLIP misclassifies some regions into false categories with similar appearances and results in incomplete segmentation masks. 

\subsection{Attention Map Fusion}
\label{3.2attnmap}
In this section, we propose an Attention Map Fusion strategy (AMF) to emphasize the global knowledge encoded by global tokens via reshaping the last-block attention. 

As shown in Figure~\ref{introduction_fig}(a), we visualize the attention maps between different patches and observe the existence of \textit{global tokens} in deeper blocks of CLIP.
The term ``global token'' means specific patches are important (\emph{i.e.,} corresponding Query-Key attention weights are super high) for all the other patches. 
These global tokens appear in the attention map as highlighted vertical lines.
Interestingly, we find that the attention pattern of global tokens aligns well with that of the \cls{} token (see the last two rows of Figure~\ref{introduction_fig}(a)), which indicates that those global tokens may encode global properties as the \cls{} token.
Though the proposed ``global token'' is similar to several concepts introduced in previous works, such as the ``global patch'' in CLIPtrase~\cite{CLIPtrase} and ``anomaly tokens'' in SC-CLIP~\cite{sc-clip}, 
we are the first to formalize global token (see Eq.~(\ref{glt})) and exploit it as an indicator of beneficial global context for segmentation in CLIP (see Eq.~(\ref{g-layer},\ref{atfuse})).

Based on the observations, we propose AMF to fuse the attention maps from early global-token emerging blocks with the last-block Query-Query attention.
Specifically, as shown in Figure~\ref{method_fig}(b), 
given a vanilla CLIP with totally $f+1$ blocks, we introduce $G(i)$ to judge whether global tokens exist in block $L_i (0\le i < f)$: 
\begin{align}
G(i) = 
\begin{cases}
1,& \text{if } \max(\prod_j \sigma \cdot A_{i, j}^{qk}) > 0\\
0,& \text{otherwise}
\end{cases} \label{glt}
,
\end{align}
where $A_{i}^{qk}$ denotes the Query-Key attention map of the $i$-th block. The $\prod_j A_{i, j}^{qk}$ means the multiplication between attention vectors for different Queries. 
Theoretically, the softmax operation ensures that all attention weights in $A$ are non-negative.
However, in practice, some of these weights can be extremely small.
When many small values are multiplied, the result may be truncated to zero because it falls below the computational precision limit of computers.
The $\sigma=100$ is set to prevent all the values from exceeding the computational precision limits.
Then we identify the block $L_{g}$ where global tokens initially emerge,
\begin{align}
g = \arg\min \{i| G(i)=1,  0\leq i < f\}. \label{g-layer}
\end{align}
We further integrate the attention weight maps of global-token emerging block $L_g$ and its following $l$ ($l<f-g$) blocks into the final Query-Query attention weight map ${A}_{f}^{qq}$ to form a new attention map ${A}_{f}$,
\begin{align}
    {A}_{f} & = \text{AMF }({A}_{g}^{qk}, ...,{A}_{g+l}^{qk},{A}_{f}^{qq}) \notag \\
    & = \frac{{A}_{g}^{qk}+ ... + {A}_{g+l}^{qk}+{A}_{f}^{qq}}{l+2} . \label{atfuse}
\end{align}
Consequently, with ${A}_{f}$, we not only enable each patch to interact with itself or its nearby patches but also allow it to aggregate image-level global properties from global tokens.
Empirically, we find that fusing with attention maps from the first and the second emerging blocks works the best, \emph{i.e.,} $l=1$.

Then our final attention output can be presented as follows: 
\begin{equation}
Z_{\text{GCLIP}} = \text{Proj}({A}_{f}\cdot v),
\end{equation}

As in different CLIP-like vision-language models (VLMs), global-token emerging layers may be different, our AMF provides a practical way to automatically identify the global-token emerging layers for various VLMs.  

\begin{figure}
  \centering
\includegraphics[width=1\linewidth]{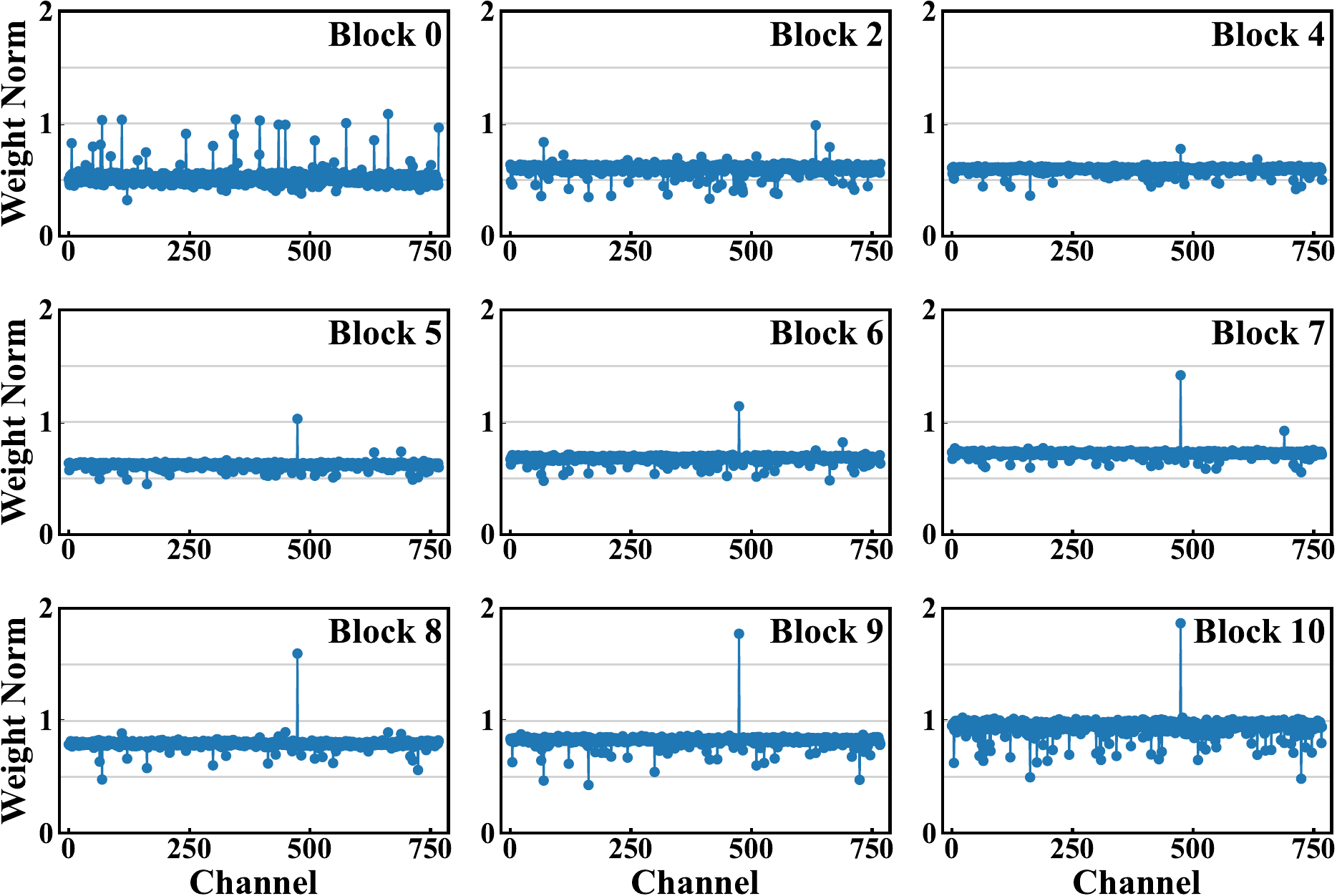}
  \caption{\textbf{Weight Norms of the second fully-connected layer in FFNs.} Starting from block 5 (CLIP ViT-B/16), we observe FFN's second fully connected layer weight norm corresponding to a specific output channel becomes unexpectedly larger than the weight norm of other channels and the average weight norm across output channels.}
\label{method_wn}
\end{figure}

\subsection{Channels Suppression}
\label{3.3wn}
In this section, we propose a Channel Suppression (CS) strategy to make last-block Value embeddings more semantically correlated.

We observe that an abnormal phenomenon exists in the weights of the second fully connected block of FFN in a Transformer block.
As illustrated in Figure~\ref{introduction_fig} (b)(2), the entropy of weight norms decreases dramatically from a certain block, and the weight norm corresponding to a specific output channel becomes unexpectedly larger than the weight norm of other channels in Figure~\ref{method_wn}.
Such an abnormal increase of specific-channel weight norm may homogeneously yield large activation of the same channel for different patch representations, which may do harm to the semantic correlations among different Value embeddings. 

Therefore, we propose a Channel Suppression strategy (CS) to make the Value embeddings of the last-block attention module more semantically correlated, as shown in Figure~\ref{method_fig}(c).
Specifically, for the weight $W \in \mathbb{R}^{D_{out} \times D_{in}}$ of the second fully connected layer of FFN in a Transformer block, we suppress the output channels which exhibit extremely high weight norms compared with the average weight norm across all output channels.


Specifically, we first calculate the weight norm $N_{d}$ for all output channels $d$ and obtain an average weight norm $\overline{{N}}$ across all output channels, \emph{i.e.,}
\begin{equation}
N_{d} = ||W_d||_{2},
\end{equation}
\begin{equation}
\overline{{N}} = \frac{\sum_{i=0}^{D_{out}-1}({N}_i)}{D_{out}},
\end{equation}
where $W_d\in\mathbb{R}^{1\times D_{in}}$.

Then, the abnormal channel $\hat{d}$ can be represented as
\begin{equation}
\hat{d} \in \{d| \frac{N_d}{\overline{{N}}}>t, d\in\{0,1,\cdots, D_{out}-1\}\},
\end{equation} 
where $t$ is a threshold to select the abnormal channels with extremely high weight norms compared with the average weight norm. Empirically, we set $t=1.5$.
Then, for each $\hat{d}$, we average the norms of all the other channels as $\overline{{N}_{\hat{d}}}$,
\emph{i.e.,}
\begin{equation}
\overline{{N}_{\hat{d}}} = \frac{\sum_{i=0,i\neq \hat{d}}^{D_{out}-1}({N}_i)}{D_{out}-1}.
\end{equation}
We retain the weights of all the other output channels while re-normalizing the weight of channel $\hat{d}$:
\begin{equation}
\hat{W}_{\hat{d}} = \varphi{(W_{\hat{d}})} =  \frac{W_{\hat{d}}}{{N}_{\hat{d}}} \times \overline{{N}_{\hat{d}}}.
\label{formula:renormalize}
\end{equation}

Suppose that an extreme decrease in the entropy of weight norms (as shown in Figure~\ref{method_wn}) occurs at block $s$,
we employ CS for each block $L_{i}$ where $s \le i \le f$. 

With the suppression, 
as shown in Figure~\ref{introduction_fig} (b)(1), we observe the Value embeddings of patches within the same semantic mask become more similar (see ``in-in'' comparison) while those from different masks become more distinct (see ``in-out'' comparison).
These results verify that CS enhances the patch-wise semantic correlation of the final Value embeddings. 



\subsection{GCLIP for training-free OVSS}
In GCLIP, both Attention Map Fusion (AMF) and Channel Suppression (CS) are employed.
With AMF, we emphasize the global knowledge encoded by global tokens by reshaping the last-block attention. 
With CS, we enhance the semantic correlation of last-block Value embeddings.
As the patch-wise visual representation is finally determined by the last-block attention and the Value embeddings, 
it is expected that GCLIP can aggregate more beneficial global knowledge into final features and yield patch-wise features with high semantic coherence for semantic segmentation.

\section{Experiments}
\label{sec:experiment}


\begin{table*}[!t]
\setlength{\tabcolsep}{7.5pt}
\renewcommand{\arraystretch}{1.7}
\centering
\caption{\textbf{Comparison with trainable open-vocabulary semantic segmentation methods (T-OVSS), unsupervised CLIP-based semantic segmentation methods (USS), and training-free open-vocabulary semantic segmentation methods (TF-OVSS).} {Among these, $^{\dag}$ means the results are obtained by running the officially released source code and $^{\ddagger}$ means the results are cited from ClearCLIP~\cite{ClearCLIP}.}}
\label{table_main}
\begin{tabular}{l|c|c|cccccccc}
\hline
 Methods & Pub. \& Year & Setting &  PASCAL VOC  & Context & \; ADE20K \; & Cityscapes & COCO Stuff & \;Avg.\;\\
    \hline
  GroupViT$^{\ddagger}$~\cite{GroupViT}& CVPR'22& \multirow{3}{*}{\makecell[c]{T-OVSS}} &$79.7$& $23.4$& $9.2$& $11.1$&$15.3$ & $27.7$\\
  CoCu~\cite{CoCu}&NeurIPS'24& & -& -& $11.1$& $15.0$&$13.6$ &-\\
  TCL~\cite{TCL}&CVPR'23 & &$77.5$ & $30.3$& $14.9$&$23.1$ &$19.6$ &$33.1$ \\
 \hline
 MaskCLIP+$^{\dag}$~\cite{MaskCLIP}& ECCV'22&\multirow{3}{*}{USS} & $70.0$ & $31.1$ & $12.2$ & $25.2$ & $19.5$ & $31.6$\\
 CLIP-S4~\cite{CLIP-S4}& CVPR'23& & $72.0$ & $33.6$ & -& -&- &-\\
 ReCLIP~\cite{ReCLIP}& CVPR'24& & $75.8$ & $33.8$ & $14.3$ & $19.9$ &$20.3$ &$32.8$\\
 \hline
 CLIP$^{\ddagger}$~\cite{CLIP}& ICML'21&\multirow{9}{*}{\makecell[c]{TF-OVSS}}& $41.8$&$9.2$ & $2.1$&$5.5$ &$4.4$ & $12.6$\\
MaskCLIP$\dag$~\cite{MaskCLIP}& ECCV'22&&$49.5$ & $21.7$ &  $9.5$ & $19.8$ & $13.6$ & $22.8$ \\
 CLIPSurgery~\cite{CLIPSurgery}& Arxiv'23&&- & -& -&$31.4$ &$21.9$ & -\\
 GEM$^{\ddagger}$~\cite{GEM} & CVPR'24&&$79.9$ & $35.9$& $15.7$&$30.8$ &$23.7$ & $37.2$\\
 SCLIP~\cite{SCLIP}& ECCV'24&&$80.4$ &$34.2$ & $16.1$& $32.2$&$22.4$ & $37.1$\\
 CLIPtrase~\cite{CLIPtrase}& ECCV'24&&$81.2$ &$34.9$ & $17.0$& -&$24.1$ & -\\
 ClearCLIP~\cite{ClearCLIP}& ECCV'24&&$80.2$$\dag$ & $35.9$&$16.7$ &$30.0$ &$23.9$ &$37.5$\\
 GCLIP (CS only) & Ours & & ${80.6}$ & ${36.2}$& ${17.8}$& ${31.3}$&${24.0}$ & $38.0$\\
 \rowcolor[gray]{.9} GCLIP (CS \& AMF) & Ours & & $\bf{81.3}$ & $\bf{36.8}$& $\bf{18.5}$& $\bf{33.7}$&$\bf{24.8}$ & $\bf{39.0}$\\
 
 \hline
\end{tabular}
\end{table*}




\begin{table}[!t]
\setlength{\tabcolsep}{6.7pt}
\renewcommand{\arraystretch}{1.7}
\caption{
\textbf{Effect of block selection for attention map fusion.} According to the results on all benchmarks, we finally fuse the attention maps of the first and the second global-token emerging blocks with the final Query-Query attention map in GCLIP.}
\centering
\begin{tabular}{c|cccccc}
\hline
$l$ &  \;VOC\; & Context & \;ADE\; & Cityscapes & \;Stuff\; & Avg.\\
\hline
0 & $81.1$ & $36.5$ & $18.3$& $32.9$ & $24.6$ & $38.7$\\
\rowcolor[gray]{.9} 1 & $81.3$ & $\bf{36.8}$ & $\bf{18.5}$ & $\bf{33.7}$& $\bf{24.8}$ & $\bf{39.0}$\\
2 & $82.0$ & $36.8$ & $18.2$ & $32.8$ & $24.8$& $38.9$\\
3 & $82.0$ & $36.6$ & $17.9$ & $31.1$ & $24.7$ & $38.5$\\
4 & $\bf{82.1}$ & $36.6$ & $18.0$ & $31.0$ & $24.6$ & $38.5$\\
\hline
\end{tabular}
\label{table_ablation_attn}
\end{table}

\begin{table}[!t]
\setlength{\tabcolsep}{4.4pt}
\renewcommand{\arraystretch}{1.5}
\centering
\caption{
\textbf{Effect of different blocks to perform channel suppression.} The block ID means we perform CS from this block to the last block. Considering average performance on all benchmarks, we choose to perform CS from block 7 to last block in GCLIP.}
\begin{tabular}{cc|cccccc}
\hline
Block &  Entropy & \;VOC\; & Context & \;ADE\;  & \;City\; & \;Stuff\; & Avg.\\
\hline
$L_5$ & $0.96$ &$78.6$ & $35.5$ & $17.3$ & $30.5$ & $23.8$& $37.1$\\
$L_6$ & $0.93$ &$79.0$ & $35.3$ & $17.3$& $30.6$ & $23.6$& $37.2$\\
\rowcolor[gray]{.9} $L_7$ &$\bf{0.53}$ & $\bf{80.6}$ & $\bf{36.2}$ & $\bf{17.8}$ & $\bf{31.3}$ & $\bf{24.0}$ & $\bf{38.0}$\\
$L_8$ & $0.28$ &$80.1$ & $36.0$ & $17.5$ & $30.4$ & $23.9$& $37.6$\\
$L_9$ & $0.09$ &$80.2$ & $35.9$ & $17.4$ & $30.5$ & $23.9$& $37.6$\\
$L_{10}$ &$0.12$ & $80.2$ & $35.9$ & $17.5$ & $30.5$ & $23.9$& $37.6$\\
\hline
\end{tabular}
\label{table_ablation_wn}
\end{table}

\subsection{Setup}
\label{4.1setup}
\noindent \textbf{Datasets}
We conduct experiments mainly on five standard benchmarks for semantic segmentation, including PASCAL VOC 2012 \cite{PASCAL_VOC}, PASCAL Context \cite{PASCAL_Context}, ADE20K \cite{ADE20K}, Cityscapes~\cite{cityscapes} and COCO Stuff~\cite{cocostuff}. 
PASCAL VOC 2012 (1,464/1,449 train/validation) contains 20 object classes, 
while PASCAL Context (4,998/5,105 train/validation) is an extension of PASCAL VOC 2010 and we treat 59 most common classes as foreground in our experiments. 
ADE20K (20,210/2,000 train/validation) is a segmentation dataset with various scenes and the 150 most common categories are considered.
Cityscapes~(2,975/500 train/validation) consists of various urban scene images of 19 categories from 50 different cities.
COCO Stuff~(118,287/5,000 train/validation) has 171 low-level thing and stuff categories excluding background class. 

\noindent \textbf{Architecture}
We use the text encoder of pre-trained CLIP~\cite{CLIP} model to generate text embeddings and modify the image encoder of CLIP to extract visual features.
For the image encoder, following general practice~\cite{ClearCLIP, SCLIP, CLIPtrase}, we adopt ViT-B/16.

\noindent \textbf{Implementation details}
Following previous works of training-free OVSS~\cite{ClearCLIP, SCLIP, CLIPtrase}, we resize the input image and employ a sliding window inference strategy. 
For inference, we only utilize category names to generate text embeddings with the prompt templates provided by CLIP~\cite{CLIP} and do not exploit further text expansions. 
We adopt CS to only modify the Value embeddings in the subsequent transformer blocks with other parts unchanged.
To make a fair comparison, we do not apply any post-processing to our evaluation results.
We employ mean intersection over union (mIoU) as the metric to evaluate our method.

\subsection{Comparison with previous state-of-the-arts}
\label{4.2comparison}

\noindent \textbf{Baseline}
We compare our method to vanilla CLIP where we take patch-wise visual features from last transformer block and compute their 
similarities with text embeddings to generate semantic masks.
Besides, we compare our method with three types of semantic segmentation methods: 
(1) Trainable methods for OVSS (T-OVSS), including GroupViT~\cite{GroupViT}, CoCu~\cite{CoCu}, and TCL~\cite{TCL};
(2) Unsupervised CLIP-based methods for semantic segmentation (USS), including MaskCLIP+~\cite{MaskCLIP}, CLIP-S4~\cite{CLIP-S4} and ReCLIP~\cite{ReCLIP};
and (3) CLIP-based methods for training-free OVSS (TF-OVSS), including CLIP~\cite{CLIP}, MaskCLIP~\cite{MaskCLIP}, CLIPSurgery~\cite{CLIPSurgery}, SCLIP~\cite{SCLIP}, GEM~\cite{GEM}, CLIPtrase~\cite{CLIPtrase} and ClearCLIP~\cite{ClearCLIP}.
For fair comparison, we choose not to compare with methods (\emph{e.g.,} ProxyCLIP~\cite{proxyclip} and {CorrCLIP~\cite{Corrclip}}) using additional large-scale pre-trained models (\emph{e.g.,} DINO~\cite{DINO}, SAM~\cite{SAM}, Stable Diffusion~\cite{SD}, \emph{etc.}) other than CLIP.
Although these methods may achieve stronger representations by incorporating multiple models, they inevitably introduce computational overhead and practical complexity.
Instead, our proposed GCLIP provides a unified framework for CLIP and CLIP-like variants with negligible extra computation.

We directly cite the corresponding results from the original papers, except that $\dag$ means the results are obtained by running the officially released source code and ${\ddagger}$ means the results are cited from ClearCLIP~\cite{ClearCLIP}.
All the numbers reported are presented as percentages.
Among these, T-OVSS methods rely on weak annotations like image-caption pairs to train the model, while USS methods rely on unlabeled images to train the model and cannot generalize to unseen classes. 
Instead, GCLIP can directly perform open-vocabulary segmentation without any training, which falls into the category of TF-OVSS.
All TF-OVSS methods are based on pre-trained CLIP with ViT-B/16 visual backbone.

\noindent \textbf{Comparison}
The comparisons with previous state-of-the-art methods on five benchmarks are demonstrated in Table~\ref{table_main}.
From Table~\ref{table_main}, we have three observations:
(1) Without training or fine-tuning CLIP, TF-OVSS methods, \emph{e.g.,} ClearCLIP, our GCLIP, \emph{etc.}, outperforms vanilla CLIP~\cite{CLIP} remarkably, which demonstrates CLIP does encode beneficial knowledge for complex visual understanding tasks.
(2) Our GCLIP even outperforms some typical T-OVSS and USS methods, showing that CLIP itself is potentially a good OVSS segmentor and our way of modifying CLIP to mine useful knowledge for segmentation is effective.
(3) Our GCLIP outperforms previous state-of-the-art TF-OVSS methods obviously, achieving new state-of-the-arts on all the five benchmarks. 
For example, on Cityscapes, GCLIP outperforms SCLIP, GEM and ClearCLIP by 1.5\%, 2.9\% and 3.7\% mIoU respectively; on ADE20K, GCLIP outperforms SCLIP, GEM, CLIPtrase and ClearCLIP by 2.4\%, 2.8\%, 1.5\% and 1.8\% mIoU.
All these results verify the effectiveness of our method of utilizing beneficial global knowledge to assist OVSS segmentation.

\subsection{Qualitative Results}
\label{4.3visualization}
\begin{figure*}
  \centering
  \includegraphics[width=1\linewidth]{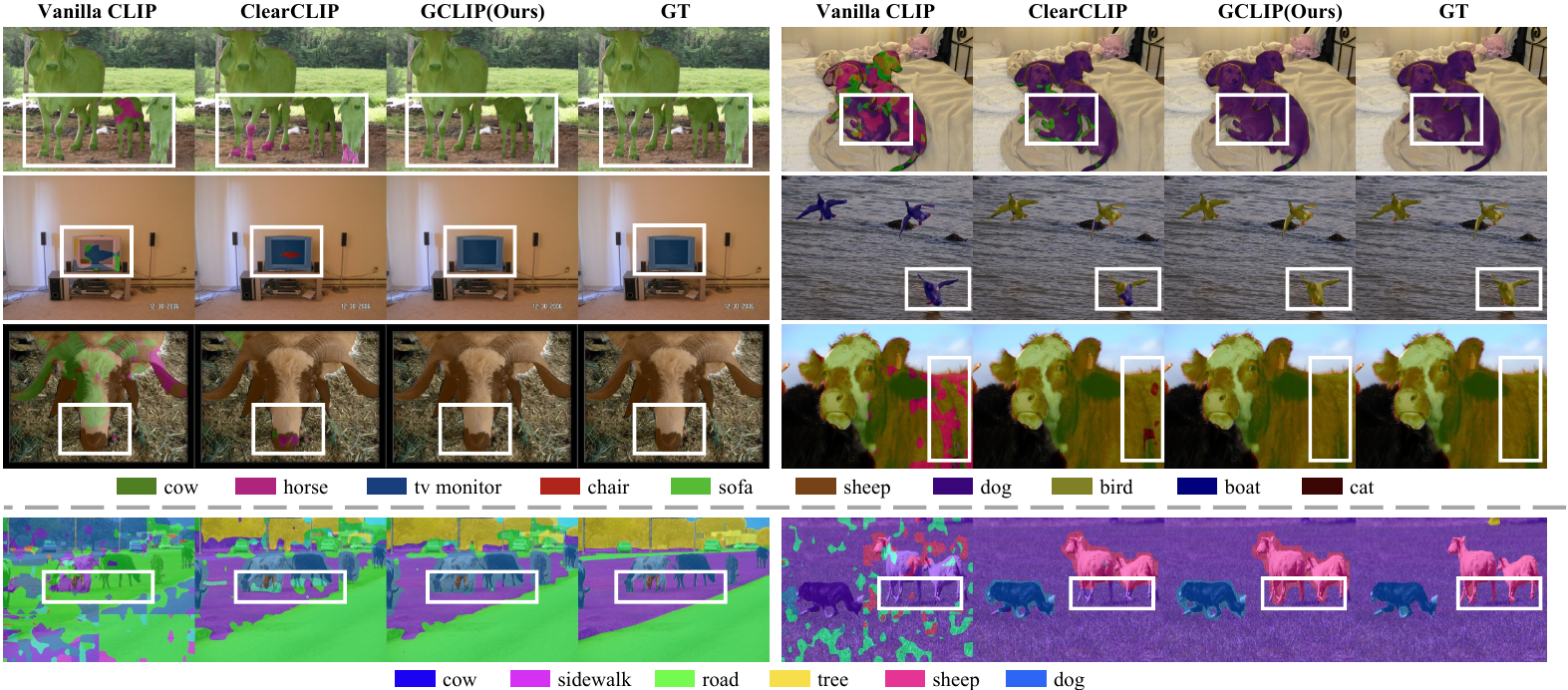}
  \caption{\textbf{Qualitative Results.} 
  We visualize the segmentation results of GCLIP on both PASCAL VOC and PASCAL Context.
  We observe that the masks generated by ClearCLIP usually fail to segment the integral target object because it may confuse semantically similar categories without sufficient global context.
  GCLIP extracts semantically correlated patch-level image features through enhancing global context information. The masks generated by GCLIP obviously outperform those of both vanilla CLIP and ClearCLIP.}
\label{visualization_fig}
\end{figure*}

We visualize the segmentation results of GCLIP on PASCAL VOC and PASCAL Context in Figure~\ref{visualization_fig}.
We observe that both ClearCLIP and GCLIP yield much better masks than vanilla CLIP. 
But the masks generated by ClearCLIP are still incomplete.
For example, when segmenting a cow (Green Mask), ClearCLIP misclassifies some regions of cow as horse (Pink Mask).
Since ClearCLIP does not fully utilize the global knowledge of CLIP, it may fail to distinguish similar yet different categories.
GCLIP avoids such confusion and yields more integral and accurate masks, through absorbing image-level global properties and enhancing the semantic correlation of Value embeddings.

\begin{table}[!t]
\setlength{\tabcolsep}{5pt}
\renewcommand{\arraystretch}{1.5}
\centering
\caption{
\textbf{Comparison with fusing \cls{} attention in AMF.} The ``\cls{} Atten.'' means we replace the patch-wise attention of global-token emerging blocks in AF  module with the attention of \cls{}. We duplicate \cls{} attention for each patch to fuse with last-block attention.} 
\begin{tabular}{c|ccccc}
\hline
Fusion & \; VOC \; & Context & \; ADE \; & Cityscapes & Stuff\\
\hline
\cls{} Atten. & $80.3$ & $35.4$ & $16.6$ & $27.0$ & $24.2$\\
\rowcolor[gray]{.9} Ours & $\bf{81.3}$ & $\bf{36.8}$ & $\bf{18.5}$ & $\bf{33.7}$ & $\bf{24.8}$\\
\hline
\end{tabular}
\label{table_ablation_clstoken}
\end{table}

\begin{table}[!t]
\setlength{\tabcolsep}{6.8pt}
\renewcommand{\arraystretch}{1.5}
\centering
\caption{
\textbf{Global tokens encode image-level global knowledge.} We exploit the classification results with \cls{} token as ground truth to evaluate the classification accuracy of global tokens.
We further provide classification accuracy with randomly-selected non-global tokens to make a comparison.
Results indicate that global tokens align well with \cls{} token in terms of encoding image-level global knowledge.}
\begin{tabular}{c|ccccc}
\hline
Tokens & \; VOC \; & Context & \; ADE \; & Cityscapes & Stuff\\
\hline
Random & $44.3$ & $36.5$ & $29.6$ & $53.1$ & $23.1$\\
Average & $51.6$ & $53.6$ & $55.7$ & $80.4$ & $47.1$\\
\rowcolor[gray]{.9} Global & $\bf{75.0}$ & $\bf{71.4}$ & $\bf{66.9}$ & $\bf{97.9}$ & $\bf{66.4}$\\
\hline
\end{tabular}
\label{table_ablation_classification}
\end{table}

\subsection{Ablation study}
\label{4.4ablation_study}

\noindent \textbf{Effect of components in GCLIP.}
As shown in Table~\ref{table_main}, we verify the effectiveness of proposed components in GCLIP.
Compared with ClearCLIP~\cite{ClearCLIP}, the channel suppression strategy (numbers with only ``CS'') on average brings 
0.5\% mIoU improvement.
Notably, it achieves 1.3\% mIoU performance gain on Cityscapes.
Additionally introducing the attention map fusion strategy (numbers of ``CS \& AMF'') yields additional 1\%  improvement. Compared to previous state-of-the-art,  GCLIP yields 1.5\% mIoU gain on average across all benchmarks.

\noindent \textbf{Effect of $l$ in attention map fusion (AMF).}
In AMF, we set $l=1$ 
in our solution, which means we fuse the Query-Key attention map of the first and the second global-token emerging blocks with the final-block Query-Query attention map.
In order to validate the effect of $l$, we perform an ablation on the effect of $l$ in AMF in Table~\ref{table_ablation_attn}.
On average, $l = 1$ yields the best results, and the performance is insensitive to $l$.


\begin{table}[!t]
\renewcommand{\arraystretch}{1.5}
\setlength{\tabcolsep}{8.5pt}
\centering
\caption{
\textbf{Effect of fusion strategy in AMF of GCLIP.} 
In this table, $L_i$ means that we begin to fuse from the $i$-th Block, while $l$ denotes that we fuse $l$ blocks following the $i$-th Block. Note that the 6-th block is the one where global token starts to emerge. Experiments are conducted on ADE20K and CS is not applied. 
}
\begin{tabular}{c|cccccc}
\hline
 &  \; $L_{0}$ \; & $L_{2}$ & \; $L_{4}$ \; & $L_{6}$ & $L_{8}$ & $L_{10}$\\
\hline
$l=0$ & $16.3$ & $17.4$ & $17.5$ & $18.0$ & $17.2$& $17.5$\\
$l=1$& $16.0$& $17.5$&$17.5$&$\bf{18.1}$&$18.0$&-\\
$l=3$& $16.5$& $17.6$&$17.8$&$17.5$&-&-\\
$l=5$& $17.2$& $17.7$&$17.7$&-&-&-\\
$l=7$& $17.6$& $17.6$&-&-&-&-\\
\hline
\end{tabular}
\label{table_ablation_amf_more}
\end{table}

\noindent \textbf{Effect of fusing strategy in AMF}
Our AMF automatically discovers the global-token emerging block $L_g$ and fuses the attention maps of block $L_g$ and $L_{g+1}$ with the final Q-Q attention.
In order to verify the effectiveness of our fusion strategy in AMF, we provide results of more fusion choices in Table~\ref{table_ablation_amf_more}.
In this table, $L_i$ means we begin to fuse from the $i$-th Block, while $l$ denotes fusing $l$ blocks following the $i$-th Block.
The ablations are performed on ADE20K.
From the table, we observe that the fusion strategy ($L_6$, $l=1$) of AMF yields the best results 
{and the optimal starting block for fusion aligns well with the discovered global-token emerging block $L_g$.}

\noindent \textbf{Effect of different blocks to perform channel suppression (CS).}
We employ CS from block 7 to the last block of CLIP in our solution, as we observe a noticeable decrease of the entropy of weight norms at these blocks (shown in Figure~\ref{introduction_fig}(c)). 
{Since the entropy of weight norms only depends on the pre-trained weights, the choice of beginning layer in CS is the same across all datasets.}
To validate the effect of this choice, 
we perform an ablation to test the effect of different blocks to perform CS in Table~\ref{table_ablation_wn}.
In this ablation, we do not include the AMF but simply test with CS.
The results show that suppressing from block 7 yields the best result on average, which is consistent with the decreasing trend of entropy of weight norms from block 7.



\noindent \textbf{Effect of GCLIP on various VLMs.}
We further test GCLIP with other typical pre-trained VLMs, including OpenCLIP~\cite{OpenCLIP} and MetaCLIP~\cite{MetaCLIP}.
Though our method is designed for CLIP, results in Table~\ref{table_ablation_backbone} show that GCLIP can be adopted to other typical VLMs and brings consistent improvement across various benchmarks. It is because we observe same issues with CLIP on those VLMs. Observation details can be found in our supplementary materials. 

{\noindent \textbf{Inference Efficiency.} 
Compared to vanilla CLIP, the additional computation cost of GCLIP during inference is introduced by fusing different attention maps. Such operations consume very little computational resource.
For example, on the whole PASCAL VOC dataset, GCLIP only adds $4.5$s of time.}

\begin{table}[!t]
\setlength{\tabcolsep}{6pt}
\renewcommand{\arraystretch}{1.5}
\centering
\caption{
\textbf{Effect of GCLIP on various VLMs.} GCLIP brings consistent improvement on various VLMs, including OpenCLIP and MetaCLIP, which further verifies the robustness of our proposed method.}
\begin{tabular}{c|c|cccc}
\hline
 Method  & VLM& \; VOC \; & \; ADE \; & \; City \;& \; Stuff \;\\
\hline
Vanilla & \multirow{3}{*}{\makecell[c]{OpenCLIP}} & $35.4$ & $2.2$ & $5.0$ & $4.3$ \\
ClearCLIP & &$78.3$& $17.4$ & $27.9$ & $23.5$\\
\rowcolor[gray]{.9}GCLIP & &$\bf{81.0}$  & $\bf{18.8}$ & $\bf{30.7}$ & $\bf{25.2}$ \\
\hline
Vanilla&\multirow{3}{*}{\makecell[c]{MetaCLIP}}& $47.2$ & $5.0$ & $5.1$ & $2.9$ \\
ClearCLIP &&$81.4$  & $18.9$ & $31.8$ & $23.1$ \\
\rowcolor[gray]{.9}GCLIP & & $\bf{83.5}$ & $\bf{18.9}$ & $\bf{32.4}$ & $\bf{23.1}$ \\
\hline
\end{tabular}
\label{table_ablation_backbone}
\end{table}

\begin{table}[!t]
\renewcommand{\arraystretch}{1.5}
\setlength{\tabcolsep}{4.8pt}
\caption{
\textbf{Comparison with training-free open-vocabulary semantic segmentation methods with ViT-L/14 backbone.} We cite the results of previous methods with ViT-L/14 backbone from ClearCLIP~\cite{ClearCLIP}.
}
\centering
\begin{tabular}{l|ccccc}
\hline
Methods &  \; VOC \; & Context & \; Cityscapes \; & ADE & Stuff\\
\hline
CLIP~\cite{CLIP} & $15.8$ & $4.5$ & $1.2$ & $2.9$ & $2.4$ \\
MaskCLIP~\cite{MaskCLIP} & $30.1$ & $12.6$ & $6.9$ & $10.1$ & $8.9$\\
SCLIP~\cite{SCLIP} & $60.3$ & $20.5$ & $7.1$ & $17.0$ & $13.1$\\
ClearCLIP~\cite{ClearCLIP} & $80.0$ & $29.6$ & $27.9$ & $15.0$ & $19.9$\\
\rowcolor[gray]{.9} GCLIP (Ours) & $\bf{82.2}$ & $\bf{31.2}$& $\bf{29.9}$ & $\bf{16.2}$ & $\bf{21.2}$\\
\hline
\end{tabular}
\label{table_sup_backbone}
\end{table}

\noindent \textbf{Quantitative results with CLIP ViT-L/14} 
In our main paper, we report our results with ViT/B-16 backbone of CLIP.
In order to verify the effectiveness of GCLIP on various backbones, we further conduct experiments on ViT-L/14 backbone of CLIP and compare GCLIP with previous CLIP-based training-free open-vocabulary semantic segmentation methods~\cite{CLIP, MaskCLIP, SCLIP, ClearCLIP} (TF-OVSS) in Table~\ref{table_sup_backbone}.
According to the results, GCLIP outperforms all previous methods for TF-OVSS with ViT-L/14 on five benchmarks.
Notably, GCLIP outperforms ClearCLIP~\cite{ClearCLIP} by 2.2\% mIoU on PASCAL VOC~\cite{PASCAL_VOC}.
The results further demonstrate the effectiveness of GCLIP with different backbones.

\noindent \textbf{Comparison with fusing \cls{} attention in AMF.}
In GCLIP, we integrate the attention from the global tokens emerging blocks into the Query-Query attention to equip the last-block attention with image-level global properties.
There exists an alternative way to duplicate the attention map of the \cls{} token and combine it with the Query-Query attention.
We then compare them in Table~\ref{table_ablation_clstoken}. 
We observe that our fusion way outperforms fusing \cls{} token.
This may be because patch-wise attention in global token emerging blocks contain more diverse global attention patterns than duplicating \cls{} attention across patches, which may avoid homogeneous visual representations while absorbing global context information. 

\noindent \textbf{Global tokens encode image-level global knowledge.}
We assume that the global tokens contain rich image-level global context. Such global context information may benefit image-level classification, similar to the effect of \cls{} token. 
In this ablation, we verify such an assumption by conducting image-level classification experiments with both global and non-global tokens in Table~\ref{table_ablation_classification}. 
First, we utilize the embedding of \cls{} token as the visual feature to perform zero-shot classification and obtain the predicted classification result for each image.
Then we use the results predicted with \cls{} token as ground truth to evaluate the zero-shot classification results with global tokens (denoted as ``Global'').
Technically, we average the classification logits of all global tokens to calculate the classification result of global tokens.
To make a comparison, we also conduct similar empirical evaluations on non-global tokens.
We either randomly select one non-global token to perform classification (denoted as ``Random'') or average classification logits of all non-global tokens to calculate the classification result (denoted as ``Average''). 
As shown in Table~\ref{table_ablation_classification}, we observe that, compared with non-global tokens, the predicted classification result of global tokens is more consistent with that of the \cls{} token, which further validates that global tokens encode rich image-level global context.

\section{Conclusion}
In this paper, we propose GCLIP for training-free open-vocabulary semantic segmentation. 
We aim to mine and utilize the global knowledge of CLIP beneficial for semantic segmentation.
We propose AMF to equip the last-block attention with global properties while not introducing homogeneous attention patterns across patches and Channel Suppression to make the Value embeddings of the last-block attention module more semantically correlated.
Via enhancing global knowledge of final features, GCLIP can generate more semantically correlated patch-level image features for TF-OVSS.
Extensive experiments demonstrate that our method achieves superior segmentation performance compared with previous state-of-the-arts. We hope our work may inspire future research to investigate how to better utilize CLIP's knowledge for complex visual understanding tasks.


\bibliographystyle{IEEEtran}
\bibliography{main}






\end{document}